%% file: main.tex
\PassOptionsToPackage{table}{xcolor}
\documentclass{fairmeta}
\usepackage{amsmath,amssymb}
\usepackage{needspace}
\usepackage{enumitem}
\usepackage{float}
\usepackage{url}
\usepackage{booktabs}
\setlist[itemize]{topsep=3pt,itemsep=2pt}
\graphicspath{{figures/}}
\hypersetup{
  pdftitle={Beyond Next-Screen Prediction: The Decision Value of GUI World Models},
  pdfauthor={Sizhe Tang, Hanhan Zhou, Jeffrey Svajlenko, Rashmi Gangadharaiah}
}
\newcommand{\takeaway}[2]{%
\begin{tcolorbox}[
  enhanced,
  colback=orange!5,
  colframe=black!65,
  boxrule=0.5pt,
  arc=2mm,
  left=2mm,right=2mm,top=1mm,bottom=1mm
]
\textbf{#1}\par\smallskip #2
\end{tcolorbox}}
\title{Beyond Next-Screen Prediction:\\ The Decision Value of GUI World Models}

\author[1,2,*]{Sizhe Tang}
\author[2]{Hanhan Zhou}
\author[2]{Jeffrey Svajlenko}
\author[2]{Rashmi Gangadharaiah}

\affiliation[1]{George Washington University}
\affiliation[2]{Amazon Web Services}

\contribution[*]{Work done during an internship at AWS AI Labs.}

\authoremails{\email{s.tang1@gwu.edu}, \email{hanhanz@amazon.com}}

\abstract{\input{files/00_abstract}}
\date{September 2026}
\begin{document}
\maketitle
\input{files/01_introduction}
\input{files/02_related_work}
\input{files/03_study_design}
\input{files/04_evaluator_intervention}
\input{files/05_direct_action_ranking}
\input{files/06_online_results}
\input{files/07_conclusion}

\clearpage
\nocite{loshchilov2019}
\bibliographystyle{assets/plainnat}
\bibliography{ref}
\clearpage
\beginappendix
\input{files/appendix_a}
\input{files/appendix_b}
\input{files/appendix_c}
\input{files/appendix_d}
\input{files/appendix_e}
\end{document}

%% file: files/00_abstract.tex
GUI world models are increasingly used to improve action selection by predicting the consequences of candidate actions, but it is unclear how much of the resulting gain comes from prediction itself rather than from the evaluator that interprets those predictions. We study this attribution problem through controlled interventions on GUI action selection. With candidates and predictions held fixed, lookahead gains can change with how the prediction is presented to the evaluator, so end-to-end gains cannot be attributed to prediction alone. We then introduce Direct Action Ranking (DAR), which learns from reference actions to score candidates without generating future screens, and compare it with a prediction-augmented variant under matched supervision. Predicted successors add no detectable ranking signal once evaluation is learned directly. Without retraining, DAR improves AndroidWorld task success by 5.8--16.4 points across three proposers, outperforms world-model lookahead, and transfers zero-shot to unseen apps, where every prediction-conditioned configuration falls below proposer confidence. These results suggest that GUI world models should be evaluated by their decision value, the incremental action-selection value of their predictions beyond a strong direct evaluator, rather than by prediction fidelity or end-to-end lookahead gain alone.

%% file: files/01_introduction.tex
\section{Introduction}

GUI agents increasingly rely on learned models of interface dynamics to reason before acting. Rather than committing immediately to the proposer's highest-ranked action, a lookahead agent can predict how the interface will change under several candidate actions and use those predicted outcomes to guide its choice. This idea has become a prominent direction in GUI agents, spanning textual outcome prediction, graphical generation, and renderable next-state simulation \citep{gu2025,chae2025wma,zheng2026}. The appeal is straightforward: anticipating action consequences can prevent mistakes before execution.

A lookahead pipeline has two components. A world model generates possible future observations; an evaluator turns them into a ranking over candidate actions. End-to-end success depends on both the information in the prediction and the evaluator's ability to use it. The same prediction can support a better choice under one evaluator and a worse choice under another. A gain from ``world-model lookahead'' is therefore a gain of the prediction--evaluator system, and it does not reveal how much came from the predicted observation.

\begin{figure}[!htbp]
\centering
\includegraphics[width=\linewidth]{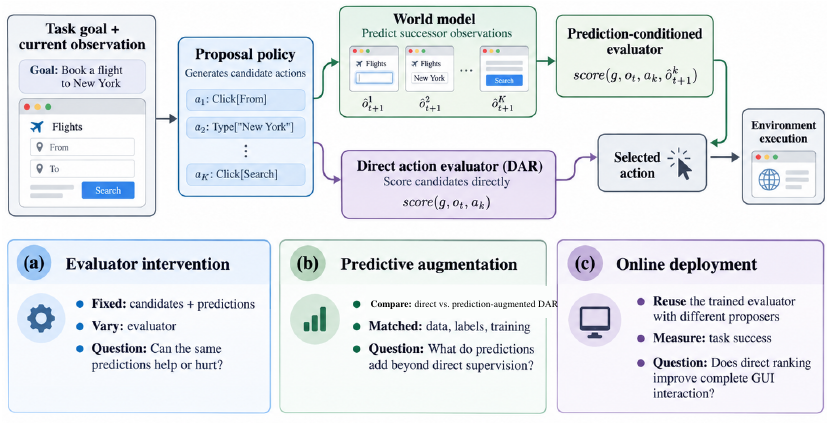}
\caption{Action selection and controlled comparisons. Lookahead and DAR use alternative paths to score the same candidate actions. We vary the evaluator with predictions fixed, compare DAR with prediction-augmented DAR under matched training, and evaluate online task completion. Interface thumbnails are schematic.}
\label{fig:1}
\end{figure}

This distinction changes what should be measured. Prediction fidelity measures whether a predicted screen resembles the true future interface; end-to-end lookahead gain measures the whole pipeline. Neither measures what the prediction contributes to the choice. We call this decision value of GUI world models: the improvement in action selection attributable to predicted observations beyond direct evaluation of the same candidates. Decision value is defined by intervention, not by fidelity. A visually accurate prediction can be decision-irrelevant when the current screen and action already determine the right choice, while a coarse prediction can still be useful if it exposes a consequence that changes which candidate is best. Measuring it therefore requires comparing direct and prediction-conditioned evaluation on the same candidates under matched supervision (Figure~\ref{fig:1}). We study this question for one-step prediction used in inference-time candidate selection.

We investigate this quantity by progressively separating prediction from evaluation. First, on AndroidControl trajectories \citep{li2024}, we keep candidate actions and cached world-model predictions fixed and change the evaluator. Identical predictions gain changes with how the prediction is presented to the evaluator, establishing evaluator choice as a first-order component of lookahead performance. We then introduce Direct Action Ranking (DAR), which learns to rank candidate actions directly from the goal and current interface using reference-action supervision, without generating a future screen. Under matched training, adding predicted successors to DAR changes reference-match AUROC by less than 0.001 across two candidate pools, isolating the incremental offline ranking signal supplied by those predictions. We also compare with proposer SFT trained on the same reference actions.

We then test DAR in complete interaction. Applied without retraining, DAR improves AndroidWorld task success by 5.8--16.4 points across three proposers, outperforms world-model lookahead. On 14 unseen Korean-language apps, DAR ranks candidates above all lookahead configurations, each of which falls below proposer confidence. Together, these results support a simple principle: a GUI world model used for action selection should be credited for the improvement its predictions add beyond a strong direct evaluator, not for the gain of the whole lookahead pipeline.

\noindent\textbf{Our contributions.}

\begin{itemize}
\item A new evaluation target for GUI world models. We formalize decision value, the improvement in action selection that predicted observations add beyond direct evaluation of the same candidates, and introduce a protocol that measures it by separating prediction from evaluation.
\item Predictions add little once evaluation is learned. Under matched supervision, predicted screens add no detectable ranking signal to a direct evaluator; with predictions fixed, lookahead gains change with how the prediction is evaluated.
\item We introduce DAR, a world-model-free selector that outperforms world-model lookahead. Without generating future screens or retraining, DAR improves task success across proposers and world models, and transfers to unseen apps and interface.
\end{itemize}

%% file: files/02_related_work.tex
\section{Background and Related Work}

\textbf{GUI agents and grounding.} SeeClick and UGround learn visual grounding \citep{cheng2024,gou2025}; SeeAct coordinates planning and grounding \citep{zheng2024}. WebVoyager and UI-TARS pursue end-to-end multimodal interaction \citep{he2024,qin2025}. These agents must both propose useful actions and identify their targets. We study the subsequent selection problem, keeping the proposer fixed while comparing ways to choose among its candidates.

\textbf{World models for GUI agents.} GUI lookahead uses textual predictions in WMA and WebDreamer \citep{chae2025wma,gu2025}, separate graphical and textual generation in ViMo \citep{luo2025}, and renderable code in Code2World and gWorld \citep{zheng2026,koh2026}. WebEvolver and WebWorld also use simulation for policy training and inference-time search \citep{fang2025,xiao2026}. \citet{xu2026} examine feedback representation and confidence in mobile guidance. End-to-end gains evaluate the complete system; our fixed-prediction and matched-training comparisons separate the effects of changing an evaluator from those of adding predicted observations.

\textbf{Learned action evaluation.} Solution verifiers \citep{cobbe2021} and process supervision \citep{lightman2024} learn to assess reasoning. Web-Shepherd learns step-level web rewards \citep{chae2025shepherd}; WebArbiter generates reasoning-based preference verdicts \citep{zhang2026}. DAR learns from reference-action matches to score GUI candidates directly. Training a prediction-augmented DAR model on the same labels then tests whether predicted successors improve this learned comparison.

%% file: files/03_study_design.tex
\section{Action Selection and Study Design}

\subsection{Proposal, Prediction, and Evaluation}

At step \(t\), a proposal policy \(\pi\) receives a goal \(g\) and observation \(o_t\) (screenshot and available interface structure) and produces candidates \(A_t=\{a_{t,1},\ldots,a_{t,K}\}\) with confidence scores \(c_{t,k}\). Policy Top-1 executes the highest-confidence candidate; every other selector reranks this fixed set. A world model predicts \(\hat{o}_{t+1}^{(k)}=W_\phi(o_t,a_{t,k})\), and an evaluator scores each candidate with or without that prediction:

\begin{equation}\label{eq:eval}
 e^{\mathrm{pred}}_{t,k}=E_\psi\!\left(g,o_t,a_{t,k},\hat{o}_{t+1}^{(k)}\right),\qquad
 e^{\mathrm{dir}}_{t,k}=E_\theta(g,o_t,a_{t,k}).
\end{equation}

The selector executes \(a_{t,\hat{k}_t}\) with \(\hat{k}_t=\arg\max_k e_{t,k}\). We use three evaluators. Lookahead uses a prompted \(E_\psi\) and falls back to Policy Top-1. DAR is a trained \(E_\theta\). Prediction-augmented DAR is a trained \(E_\psi\) that matches DAR in candidates, labels, backbone, and objective, and additionally receives \(\hat{o}_{t+1}^{(k)}\).

\subsection{Reference Labels and Metrics}

For a recorded reference action \(a_t^{\mathrm{ref}}\), a domain-specific matcher defines

\begin{equation}\label{eq:labels}
 y_{t,k}=\mathbb{1}[a_{t,k}\simeq a^{\mathrm{ref}}_t].
\end{equation}

Several candidates can match. Each metric answers a different question. Reference Coverage@K: does any candidate match? Reference-Match AUROC: pooled over decisions, do matching candidates score higher than non-matching ones? Reference-Match Accuracy: does the selected action match, $y_{t,\hat{k}_t}=1$? Task Success Rate: does the agent complete the task, as verified by the benchmark? Together they trace a gain from available candidates to ranking quality to task completion.

\subsection{Decision Value and Comparisons}

\textbf{Decision value.} For a metric $M$, the decision value of $W_\phi$ relative to a direct evaluator $E_\theta$ is

\begin{equation}\label{eq:dv}
 \mathrm{DV}_M(W_\phi\mid E_\theta)=M(E_\psi;W_\phi)-M(E_\theta).
\end{equation}

Here $E_\psi$ is trained identically to $E_\theta$ and differs only in receiving $\hat{o}_{t+1}^{(k)}$. DV is defined relative to a direct evaluator, not for the world model alone: the same predictions may add value to a weak direct evaluator and none to a strong one.

\textbf{Comparisons.} We use three comparisons (Figure~\ref{fig:1}). Evaluator intervention (Section 4) holds candidates and predictions fixed and varies a prompted $E_\psi$, testing whether the evaluator alone changes the effect of lookahead. Prediction augmentation (Section 5) estimates DV with the matched DAR/prediction-augmented DAR pair, using Reference-Match AUROC as $M$. Online deployment (Section 6) compares Policy Top-1, Lookahead, and DAR by task success; trajectories may diverge as different actions are selected. Lookahead does not meet the matching condition, so this comparison does not measure DV; it shows how a prompted prediction pipeline compares with a trained direct evaluator. Section 5.4 also compares DAR with proposer SFT trained on the same reference actions. All comparisons study one-step prediction for inference-time candidate selection.

\begin{table}[!htbp]
\centering
\caption{Diagnostic signals for useful overrides. Override-Benefit AUROC on each evaluator's informative subset. Each evaluator has a selected-minus-Top-1 contrast column and a per-candidate scalar column; set- and step-level scores span both columns. The first four rows require the reference successor; the remaining signals are available before execution. Darker/lighter lavender marks the highest/second-highest scores per evaluator and availability block. Full statistics: Appendix A.3.}
\includegraphics[width=\linewidth]{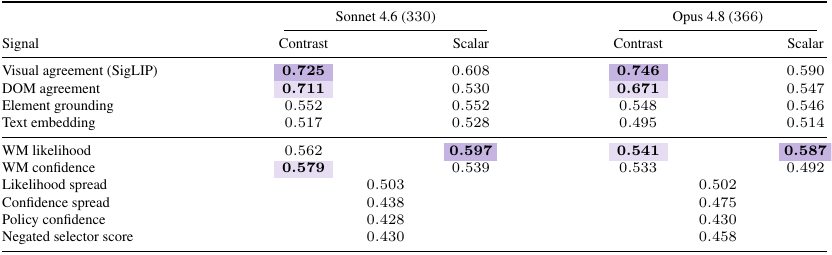}
\label{tab:1}
\end{table}

\textbf{Data.} AndroidControl \citep{li2024} provides the offline trajectories: a 3,000-step episode-level evaluation split and separate selector and SFT training pools. AndroidWorld \citep{rawles2025} evaluates complete tasks.

%% file: files/04_evaluator_intervention.tex
\section{Separating Prediction from Action Evaluation}

\subsection{Fixed Predictions, Different Decisions}

We first hold candidates and predictions fixed and vary how they are evaluated. On 3,000 AndroidControl evaluation steps, we use Qwen2.5-VL-7B candidates \citep{bai2025qwen25} ($K\leq 3$) and cached Code2World-8B predictions \citep{zheng2026}. These candidate sets leave room to improve selection without changing the proposer: 56.1\% contain a reference match, while Policy Top-1 matches on 43.8\% (Table~\ref{tab:2}).

With these fixed predictions, lookahead raises Reference-Match Accuracy over Policy Top-1 by 1.7 points with Claude Sonnet 4.6 and 2.5 points with Claude Opus 4.8; both 95\% episode-clustered intervals exclude zero (Appendix A.1). The gain also depends on how the prediction reaches the evaluator: with Opus 4.8 and the same predicted successors, re-encoding them as an accessibility tree or plain text lowers the point estimate from +2.5 to +1.2 points (Appendix D). End-to-end lookahead performance therefore characterizes a prediction--evaluator pair, not the prediction alone. This comparison does not measure decision value, which requires matched evaluators (Section 3.3). Figure~\ref{fig:2} shows one decision in which Top-1 and lookahead choose different clicks, while the direct evaluator of Section 5 selects the reference scroll; Appendix E gives further examples.

\subsection{Which Signals Identify Useful Overrides?}

We next ask whether any signal separates useful overrides from harmful ones. For $\Delta_t=y_{t,\hat{k}_t}-y_{t,k_t^{\mathrm{top}}}$, Override-Benefit AUROC distinguishes overrides that recover a reference match ($\Delta_t=+1$) from those that lose one ($\Delta_t=-1$), computed on each evaluator\'s decisions with $\Delta_t\neq 0$ (Table~\ref{tab:1}).

The strongest signals all require the reference successor, which is unavailable at decision time. Visual (SigLIP; \citealp{zhai2023}) and DOM agreement with that successor separate useful overrides best, and comparing the selected candidate with Top-1 is more informative than scoring it alone, for all evaluators. Signals available before execution are weaker and inconsistent across evaluators: world-model confidence contrast is strongest under self-evaluation, likelihood spread stays at chance, and confidence spread falls below 0.5 for two evaluators. No pre-execution signal is consistently informative, which motivates learning the candidate comparison directly from reference actions (Section 5). Appendix D analyzes lookahead depth and prediction representation.

\takeaway{Takeaway 1: Lookahead gains belong to the prediction--evaluator pair.}{With predictions fixed, the gain changes with how the prediction is presented to the evaluator.}

%% file: files/05_direct_action_ranking.tex
\section{Learning to Select Actions Directly}

\subsection{DAR: Reference-Supervised Action Ranking}

DAR instantiates the direct evaluator $E_\theta$ in Eq. 1: it is both the control for measuring decision value and a practical selector. It scores each candidate from the goal, current interface, and action description, using reference actions only to construct training labels. The mobile input includes the screenshot and available accessibility elements with bounding boxes. We adapt Qwen3-VL-8B-Instruct \citep{bai2025qwen3} with frozen base weights, LoRA adapters \citep{hu2022}, and a linear scoring head on the final-token representation. Each candidate is scored independently; candidates from the same observation are grouped when computing the ranking loss.

The listwise term \citep{cao2007} trains the within-set ordering needed for selection, while the pointwise term supervises whether each candidate matches the reference. For \(P_t=\{k:y_{t,k}=1\}\neq\emptyset\),

\begin{equation}\label{eq:loss}
\begin{aligned}
\mathcal{L}_{\mathrm{DAR}}&=\mathcal{L}_{\mathrm{rank}}+\beta\mathcal{L}_{\mathrm{match}},\\
\mathcal{L}_{\mathrm{rank}}&=-\frac{1}{|P_t|}\sum_{k\in P_t}\log\frac{\exp(e_{t,k})}{\sum_j\exp(e_{t,j})},\\
\mathcal{L}_{\mathrm{match}}&=\frac{1}{K}\sum_k\operatorname{BCE}(\sigma(e_{t,k}),y_{t,k}).
\end{aligned}
\end{equation}

For all-negative groups, $\mathcal{L}_{\mathrm{rank}}=0$ and pointwise supervision remains. The listwise term targets within-state competition, while the pointwise term anchors reference compatibility and still trains all-negative groups; Section 6.4 tests the online effect of this combination. We average over intact candidate groups, use $\beta$ = 0.5, train for two epochs, and select the highest-scoring candidate. Appendix B.1 gives filtering, input construction, domain-specific matching, and optimization settings.

\subsection{What Direct Supervision Achieves}

Reference supervision improves candidate discrimination well beyond proposer confidence (Table~\ref{tab:2}). On the shared AndroidControl test pool, DAR trained on either candidate source raises AUROC by 0.17--0.18 over policy confidence. A separate Mind2Web-trained evaluator \citep{deng2023} also improves on confidence across all three splits, including the site and domain shifts.

\begin{table}[!htbp]
\centering
\caption{Offline reference-match ranking (AUROC). Both AndroidControl columns use one Qwen2.5-VL-7B test-candidate pool; column labels identify DAR's training-candidate source. Mind2Web uses a separate web evaluator. Darker/lighter lavender marks the highest and second-highest scores within each column.}
\includegraphics[width=\linewidth]{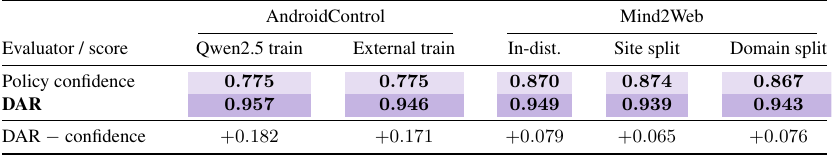}
\label{tab:2}
\end{table}

Better discrimination also produces better choices. On the original $K\leq 3$ pool, DAR raises selection accuracy from 43.8\% to 50.1\% (Table~\ref{tab:3}(a)). Adding a second proposer decode raises both reference coverage and DAR accuracy but widens the gap between them (Policy Top-1 uses the first decode throughout): more matching actions are available, yet the evaluator still misses some. A matching action must be both proposed and selected, so proposal coverage and candidate evaluation are complementary.

More supervision improves the learned comparison itself. Expanding a nested training pool raises AUROC from 0.900 to 0.957 and improves selection conditional on a matching candidate (Table~\ref{tab:3}(b)).

\begin{table}[!htbp]
\centering
\caption{Candidate budget, selector supervision, and an SFT baseline. 3,000 AndroidControl decisions. (a) Candidate budget for a fixed proposer. (b) Nested DAR training sets; conditional match accuracy is computed on decisions whose candidate set contains a reference match. (c) Qwen3-VL-32B proposer SFT, evaluated on the same evaluation fold; |A| is the deduplicated candidate count. The two-decode setting has mean candidate count 3.64. Full settings: Appendices B.2--B.4.}
\includegraphics[width=\linewidth]{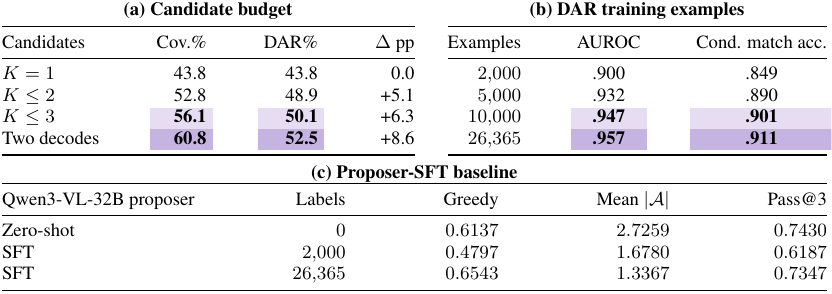}
\label{tab:3}
\end{table}

\begin{figure}[!htbp]
\centering
\includegraphics[width=\linewidth]{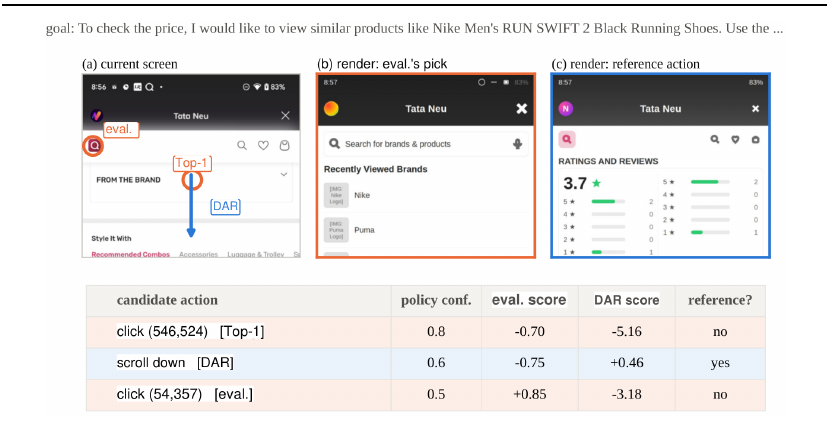}
\caption{Different choices on a recorded AndroidControl step. To find product prices, Top-1 and lookahead choose different clicks; DAR selects the reference scroll. Panels show the current screen and predictions for the lookahead and reference actions. The table gives candidate scores; ``eval.'' denotes the lookahead evaluator. Case selection: Appendix E.}
\label{fig:2}
\end{figure}

All settings use the same two-epoch schedule. Appendices B.3--B.4 give the full training and candidate-generation settings.

\subsection{What Do Predicted Observations Add?}

We now measure decision value directly: do predictions improve an evaluator that has already learned from reference actions? We train DAR and a prediction-augmented DAR model independently on an external-proposer candidate pool and a Qwen2.5-VL-7B-proposed pool with the same backbone, candidates, labels, objective, and optimization; the predicted successor is the only additional input to prediction-augmented DAR. Adding predictions changes AUROC by $-$0.0003 and +0.0005, with 95\% intervals within $\pm$0.004 (Appendix B.1), whereas direct supervision raises AUROC by 0.17--0.18 over policy confidence. Once evaluation is learned directly, predicted screens add no detectable ranking signal. Section 4.1 showed that lookahead gains depend on how predictions are evaluated; this comparison shows that, for a learned evaluator, the predictions themselves add little. Because pooled AUROC and online success can diverge (Section 6.4), this result bounds the offline ranking signal of predictions but does not by itself rule out online decision value; Section 6 tests the choices a learned selector makes online.

\takeaway{Takeaway 2: Predictions add no detectable ranking signal once evaluation is learned.}{Under matched supervision, predicted successors change reference-match AUROC by at most 0.004 (95\% CI), against a 0.17--0.18 gain from direct supervision.}

\subsection{Comparison With Proposer SFT}

A natural alternative is to spend the same reference actions on the proposer. Full-budget Qwen3-VL-32B SFT raises greedy single-step matching from 61.4\% to 65.4\% but shrinks the candidate set (mean deduplicated size from 2.73 to 1.34). DAR still adds 2.7 points on the SFT candidates, so training the proposer and learning to evaluate its candidates are complementary. With 2,000 labels, SFT falls below its zero-shot baseline, whereas DAR trained on 2,000 labels improves its own candidate pool by 3.8 points; the two settings differ in proposer and candidate pool, so this contrast is only qualitative. Appendix B.2 gives the full protocol and additional results.

%% file: files/06_online_results.tex
\section{From Action Ranking to Task Success}

\subsection{Online Selection on AndroidWorld}

Offline reference matching evaluates individual decisions; an online agent must complete a sequence of them. We train a mobile DAR model once, on Qwen2.5-VL-7B-proposed AndroidControl candidates, and apply it without retraining to three proposers on 116 AndroidWorld tasks \citep{rawles2025}, with three seeds and up to three candidates per decision. Lookahead scores Code2World-8B predictions with either an independent evaluator or the proposer itself (Self), and retains the confidence-based fallback. Task and candidate-generation protocols are shared within each proposer (Appendix C.1). This setting compares a trained direct evaluator with prompted lookahead; it does not measure decision value (Section 3.3).

\begin{table}[!htbp]
\centering
\caption{AndroidWorld application-family results (\%). 116 tasks, three seeds. All: mean $\pm$ s.e.m. External/Self use Code2World. Counts appear below family names; shading ranks selectors per proposer and column. Full paired statistics: Appendix C.1.}
\includegraphics[width=\linewidth]{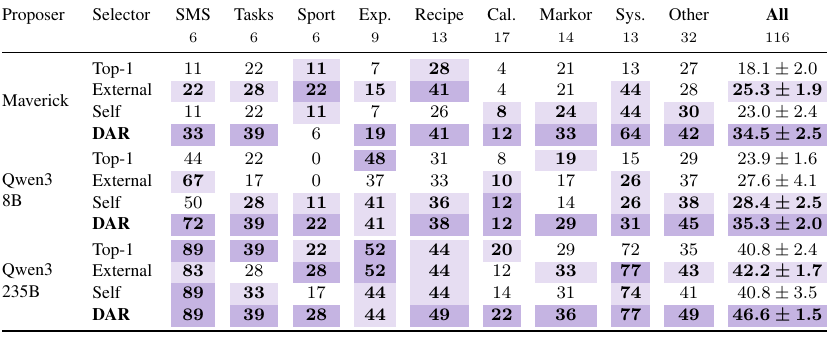}
\label{tab:4}
\end{table}

DAR has the highest mean task success for every proposer (Table~\ref{tab:4}). It improves over Top-1 by 16.4, 11.5, and 5.8 points and over the two lookahead configurations by 4.3--11.5 points; all nine paired contrasts have $p<0.05$ (exact McNemar, uncorrected; Table~\ref{tab:14}), the smallest being +4.3 over External with Qwen3-VL-235B ($p=0.049$). These gains hold under a candidate-generation shift: DAR was trained only on Qwen2.5-VL-7B proposals, yet ranks actions from Llama-4 Maverick and two Qwen3-VL models without retraining or next-screen generation. As Top-1 success rises from 18.1\% to 40.8\%, DAR's gain falls from 16.4 to 5.8 points, consistent with less selection headroom for stronger proposers.

\subsection{Changing the World Model}

We replace Code2World-8B with gWorld-8B \citep{koh2026} while keeping the proposer and selector protocols fixed. Each world model uses its native input format: Code2World receives goal conditioning and an action annotation, whereas gWorld uses a raw screenshot without a goal slot.

\begin{table}[!htbp]
\centering
\caption{AndroidWorld with gWorld-8B (\%). Same 116 tasks and families as Table~\ref{tab:4}; entries are three-seed means. $\Delta$ is the difference between the displayed three-seed mean success rates (gWorld--Code2World). Top-1/DAR do not use a world model.}
\includegraphics[width=\linewidth]{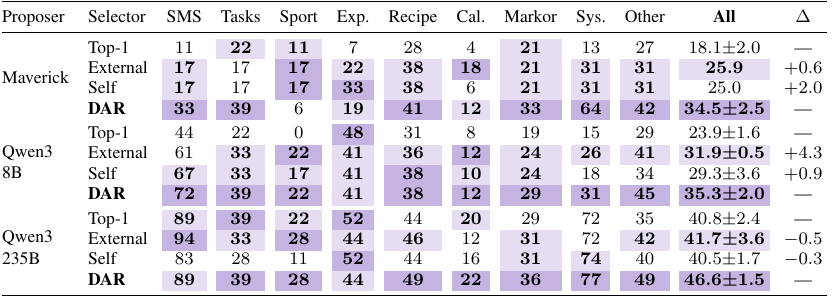}
\label{tab:5}
\end{table}

The swap changes lookahead success by $-$0.5 to +4.3 points depending on the proposer, while DAR, which uses no world model, keeps the highest mean for every proposer (Table~\ref{tab:5}). The predictor changes the evidence lookahead receives; direct evaluation provides a reference point that does not move with it.

\subsection{Transfer to Korean Apps and Web Tasks}

\textbf{New applications and language.} On MWMBench-KApps \citep{koh2026}, we apply the same mobile DAR model without retraining to 14 unseen Korean-language apps. This changes the application inventory and interface language, and two of the three proposers were not seen in training. DAR has the highest AUROC for every proposer (Table~\ref{tab:6}). Every prediction-conditioned configuration, with either world model and either evaluator, scores below policy confidence: under this shift, conditioning on predicted screens ranks candidates worse than the proposer's own confidence. Details are in Appendix C.2.

\begin{table}[!htbp]
\centering
\caption{KApps transfer. Reference-Match AUROC (mean $\pm$ SD over three seeds) across 14 Korean apps. The same DAR evaluator is used without retraining across proposers; shading ranks each column.}
\includegraphics[width=\linewidth]{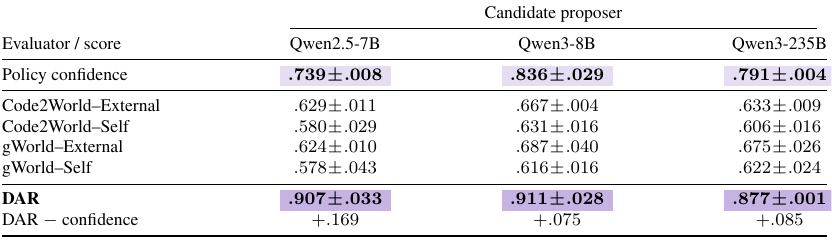}
\label{tab:6}
\end{table}

\textbf{Web interaction.} For web tasks, we train a separate DAR evaluator on Mind2Web and deploy it on 98 MiniWob tasks with three seeds \citep{deng2023,liu2018}. MiniWob changes the interaction substrate as well as the task distribution, so we use a web-specific evaluator rather than transfer the mobile one. Here the ordering depends on the proposer (Table~\ref{tab:7}): lookahead (External) outperforms DAR with Qwen3-VL-8B (38.9\% vs. 32.0\%), whereas DAR outperforms it with Qwen3-VL-235B (44.4\% vs. 40.8\%). Direct evaluation therefore does not win everywhere.

Because the two selectors differ in both evaluator and input, this comparison does not isolate the contribution of the prediction.

\begin{table}[!htbp]
\centering
\caption{MiniWob task-family results (\%). 98 tasks $\times$ 3 seeds. All: mean $\pm$ s.e.m.; parentheses: paired DAR--Top-1 gain $\pm$ s.e.m. Shading ranks each proposer and column.}
\includegraphics[width=\linewidth]{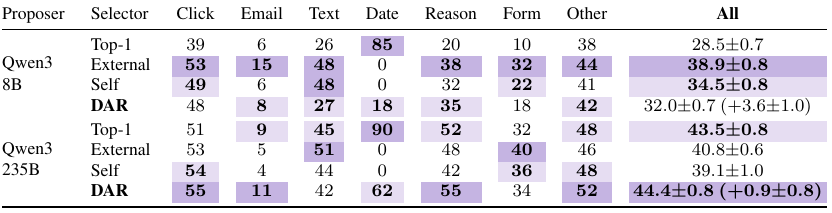}
\label{tab:7}
\end{table}

\subsection{When Offline Ranking and Task Success Diverge}

Offline ranking does not always predict online success. Removing the pointwise loss lowers AUROC from 0.957 to 0.934, but reduces conditional match accuracy from 0.911 to 0.890 and task success by 6.0 points, with worse calibration (Table~\ref{tab:8}(b); Appendix C.4). The backbone variants point the same way on a smaller scale: GLM has slightly higher AUROC than MiMo but lower mean task success, and both improve on Top-1 (Table~\ref{tab:8}(a)).

Pooled AUROC measures positive--negative orderings across candidates, whereas execution depends on the winner within each candidate set. Similar AUROC can therefore hide different action choices, and task success depends on which within-set decisions change. This is why the matched comparison in Section 5.3 bounds only the offline ranking signal of predictions, while AndroidWorld measures the downstream value of a selector's choices.

\begin{table}[!htbp]
\centering
\caption{Evaluator and objective sensitivity. AndroidControl offline; AndroidWorld online (116 tasks, Qwen3-VL-8B, three-seed mean $\pm$ s.e.m.). Conditional match accuracy is computed on decisions whose candidate set contains a reference match; full statistics are in Appendix C.4.}
\includegraphics[width=\linewidth]{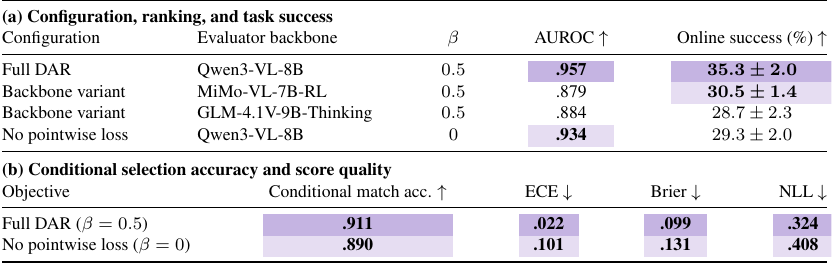}
\label{tab:8}
\end{table}

\takeaway{Takeaway 3: A trained direct evaluator sets a high bar for lookahead.}{Without retraining, DAR has the highest mean success on AndroidWorld and the highest AUROC on KApps for every proposer; DAR also improves Top-1 on both MiniWob proposers.}

%% file: files/07_conclusion.tex
\section{Conclusion}

GUI lookahead couples prediction with evaluation, so end-to-end gains cannot be attributed to prediction alone. We propose measuring a GUI world model by its decision value: what its predictions add beyond a direct evaluator of the same candidates. Under matched supervision, predictions add no detectable ranking signal once evaluation is learned, and the direct evaluator, DAR, outperforms lookahead online without generating future screens. Our study covers one-step prediction for inference-time selection; future GUI world models should report decision value alongside prediction fidelity and end-to-end gain.

%% file: files/appendix_a.tex
\section{Supporting Details for the Fixed-Prediction Study}

\subsection{Candidate Pool and Evaluator Configurations}

All evaluator comparisons use the same 3,000 AndroidControl evaluation steps, Qwen2.5-VL-7B candidates ($K\leq 3$), and cached Code2World-8B predictions. Reference matches follow the benchmark's deterministic element--bounding-box matcher. The self-evaluator is Qwen2.5-VL-7B. The external evaluators are Claude Sonnet 4.6 (us.anthropic.claude-sonnet-4-6) and Claude Opus 4.8 (us.anthropic.claude-opus-4-8).

The paired changes in Reference-Match Accuracy relative to Policy Top-1 are $-$1.4 percentage points for self-evaluation (95\% interval: [$-$2.3, $-$0.6]), +1.7 for Claude Sonnet 4.6 ([+0.6, +2.9]), and +2.5 for Claude Opus 4.8 ([+1.3, +3.8]). Intervals use episode-clustered bootstrap resampling \citep{efron1979}.

\subsection{Override Filtering and Signal Interpretation}

Using the reference-match labels from Eq. (2), define

\begin{equation}\label{eq:override}
\Delta_t=y_{t,\hat{k}_t}-y_{t,k_t^{\mathrm{top}}}.
\end{equation}

Here $\hat{k}_t$ and $k_t^{\mathrm{top}}$ index the lookahead and Policy Top-1 choices. Override-Benefit AUROC distinguishes $\Delta_t=+1$ from $\Delta_t=-1$ on decisions with $\Delta_t\neq 0$; its labels therefore measure whether an override improves or harms reference matching.

For each evaluator, Override-Benefit AUROC is computed on decisions where that evaluator's override changes reference-match status. Claude Opus 4.8 changes the selected action on 797 of the 3,000 decisions, of which 366 have $\Delta_t\neq 0$; Qwen self-evaluation and Claude Sonnet 4.6 yield 172 and 330 informative decisions, respectively. The confidence-based short circuit remains part of the lookahead protocol throughout.

Policy confidences are verbalized scores with a few coarse levels. We preserve each signal's direction when reporting AUROC, so values below 0.5 indicate association in the opposite direction. Reference-grounded signals compare a predicted observation with the successor recorded for the reference action. Contrast scores subtract agreement for the Policy Top-1 candidate from agreement for the lookahead-selected candidate; per-candidate scores omit that subtraction, and spread signals summarize variation across the candidate set.

\subsection{Diagnostic Signals}

Table~\ref{tab:9} reports the full signal sweep with episode-clustered confidence intervals. Benjamini--Hochberg correction is applied within each evaluator panel \citep{benjamini1995}.

\begin{table}[!htbp]
\centering
\caption{Complete diagnostic signal sweep. Directional Override-Benefit AUROC with 95\% episode-clustered bootstrap intervals. Each column uses its evaluator's own informative overrides: Sonnet 4.6 $n=330$ and Opus 4.8 $n=366$. AUROC values with within-panel Benjamini--Hochberg adjusted $q<0.05$ are bold. ``Contrast'' denotes selected-minus-Policy Top-1; ``scalar'' denotes a per-candidate or step-level score.}
\includegraphics[width=\linewidth]{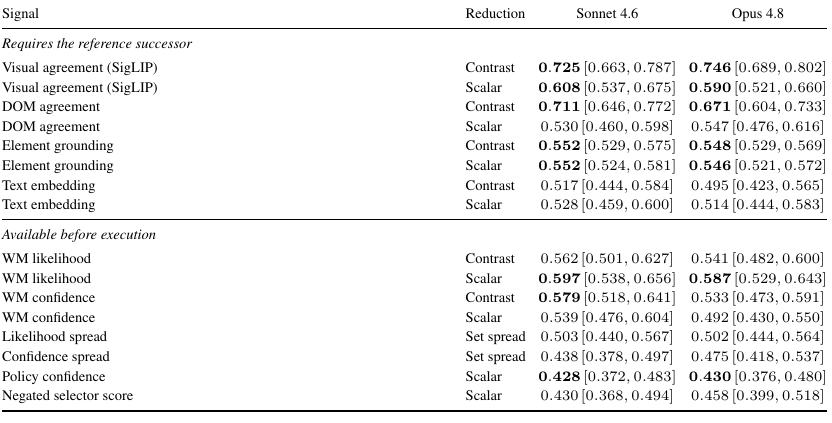}
\label{tab:9}
\end{table}

%% file: files/appendix_b.tex
\section{Direct Action Ranking: Implementation and Sensitivity}

\subsection{Implementation and Evaluation Conventions}

DAR evaluates one candidate at a time. The reference action and the other candidates are not model inputs; candidates from the same observation are grouped only to compute the listwise training loss. The mobile input contains the current screenshot, goal, visible accessibility elements with bounding boxes, and the proposed action with its resolved target element. The web instantiation uses the goal and interactable-element representation with the proposed element and operation. Mobile labels follow the deterministic AndroidControl element--bounding-box matcher \citep{li2024}; web labels use element, operation, and value equivalence.

Table~\ref{tab:10} gives the main mobile DAR training configuration. Groups in which every candidate matches are dropped during training. For retained groups with no match, the listwise term is zero and the pointwise loss supplies negative supervision. The same mobile scoring prompt is used for training and evaluation.

The two AndroidControl columns of Table~\ref{tab:2} use the same Qwen2.5-VL-7B-proposed evaluation candidate pool; the two DAR evaluators are trained on Qwen2.5-VL-7B and external-proposer candidates, respectively. The Mind2Web columns use a web DAR evaluator trained on Mind2Web and evaluated on the in-distribution, site, and domain splits \citep{deng2023}.

\textbf{Matched predictive augmentation.} Section 5.3 evaluates predictive augmentation on an external-proposer pool and a Qwen2.5-VL-7B-proposed candidate pool. Within each pool, DAR and prediction-augmented DAR use matched candidates, labels, backbone, objective, and optimization settings; the latter additionally receives the predicted successor. The paired AUROC changes are $-$0.0003 (95\% CI: [$-$0.003, +0.003]) and +0.0005 (95\% CI: [$-$0.003, +0.004]), respectively, using 2,000 episode-clustered bootstrap resamples.

\begin{table}[!htbp]
\centering
\caption{Main mobile DAR configuration.}
\includegraphics[width=\linewidth]{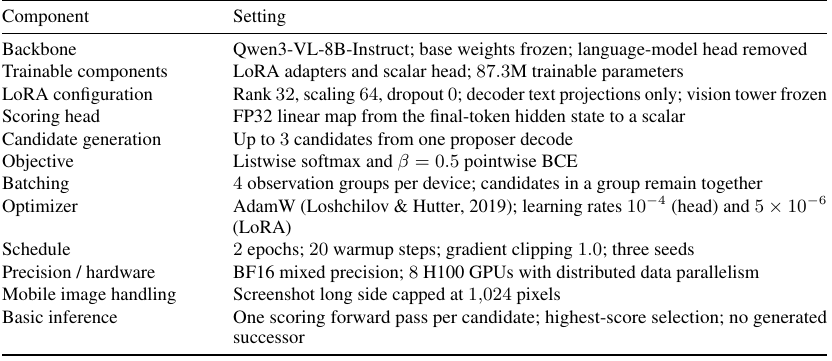}
\label{tab:10}
\end{table}

\subsection{Policy-SFT Comparison Baseline}

The policy-SFT baseline uses Qwen3-VL-32B-Instruct with LoRA rank 64 and scaling 128 on decoder text projections; the vision tower is frozen. The full-budget SFT setting uses the same 26,365 reference-action pool, prompt skeleton, and action vocabulary as the full-budget DAR comparison. All outputs are evaluated in the same pixel-coordinate action space: the zero-shot checkpoint natively emits Qwen3-VL's 0--1000 coordinates, which are deterministically mapped to pixels before action deduplication, while the SFT checkpoints emit pixel coordinates directly.

\begin{table}[!htbp]
\centering
\caption{Proposer-SFT baseline. Greedy reference matching, candidate-set size, and Pass@3 on the shared evaluation fold.}
\includegraphics[width=\linewidth]{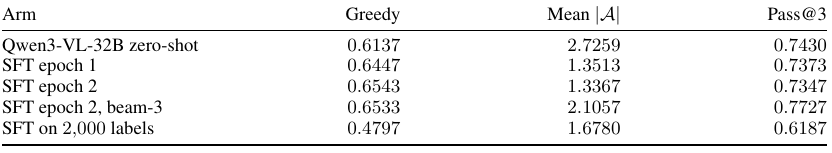}
\label{tab:11}
\end{table}

On the SFT epoch-2 candidate sets, DAR adds 0.0267 selected-action matching. With 2,000 labels, the DAR selector improves selected-action matching by 0.038 on its fixed candidate pool.

\subsection{Candidate-Set Size}

Table~\ref{tab:12} evaluates all 3,000 AndroidControl evaluation decisions. The $K\leq 3$ settings are nested subsets of the original pool: the Policy Top-1 candidate is retained, and lower-confidence alternatives are removed first. The two-decode setting adds an independent second decode of the same prompt and deduplicates actions by equivalence. Policy Top-1 remains the highest-confidence candidate from the first decode throughout.

\begin{table}[!htbp]
\centering
\caption{Candidate-budget sensitivity. Reference coverage is the proportion of decisions containing at least one matching candidate. Selected-action columns report Reference-Match Accuracy; gains are absolute fractions. The two-decode setting has mean candidate count 3.64.}
\includegraphics[width=\linewidth]{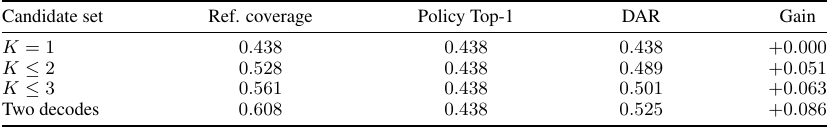}
\label{tab:12}
\end{table}

DAR improves selected-action matching over Policy Top-1 whenever alternative candidates are available. The second decode increases both reference coverage and selection accuracy.

\subsection{Training-Data Sensitivity}

Table~\ref{tab:13} uses nested subsets of one Qwen2.5-VL-7B-proposed training pool with the configuration in Table~\ref{tab:10}. All models are evaluated on the same 3,000-decision pool. Training examples count sampled observations before the all-matching filter; groups count the observations retained after that filter.

\begin{table}[!htbp]
\centering
\caption{Training-data sensitivity. AUROC includes 95\% intervals. Conditional match accuracy measures selected-action matching on observations whose candidate set contains a reference match.}
\includegraphics[width=\linewidth]{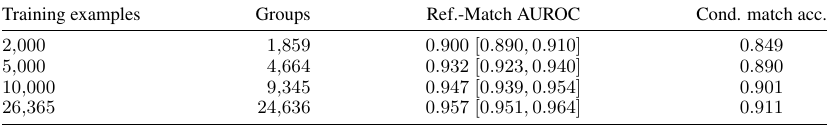}
\label{tab:13}
\end{table}

Both candidate discrimination and conditional selection accuracy improve as the training set grows.

%% file: files/appendix_c.tex
\section{Online Evaluation, Transfer, and Sensitivity}

\subsection{AndroidWorld Deployment and Paired Comparisons}

The mobile DAR model is trained on Qwen2.5-VL-7B-proposed AndroidControl candidates and reused for all three AndroidWorld proposers without parameter updates. Each task is evaluated under seeds \{30, 31, 32\}, with up to three candidates from a single proposer decode at each decision. Policy Top-1 selects by proposer confidence; DAR selects by its learned score. Lookahead--External and Lookahead--Self score Code2World-8B successors with fixed evaluator prompts across all online runs; both retain the confidence-based short circuit to Policy Top-1.

All 116 tasks remain in the denominator, and execution errors count as failures for the affected configuration. Table~\ref{tab:4} gives the application-family breakdown. Table~\ref{tab:14} reports paired contrasts using 2,000 task-clustered bootstrap resamples \citep{efron1979} and exact McNemar tests \citep{mcnemar1947}. Online trajectories may diverge after different selected actions, so pairing is defined by task and seed rather than by later candidate sets.

\begin{table}[!htbp]
\centering
\caption{Paired AndroidWorld task-success contrasts. DAR minus each baseline, pooled over seeds 30, 31, and 32. Gains are within-seed paired differences in percentage points. Intervals use task-clustered bootstrap; p-values use exact McNemar tests. Paired differences are computed before rounding marginal means.}
\includegraphics[width=\linewidth]{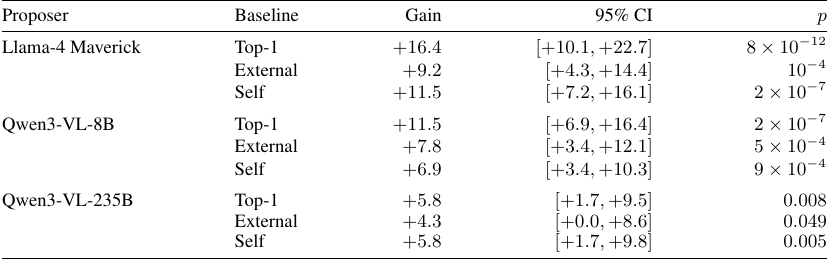}
\label{tab:14}
\end{table}

\subsection{MWMBench-KApps Transfer}

KApps contains 488 steps from 230 episodes across 14 Korean applications. The same mobile DAR evaluator is applied without retraining to candidates from all three proposers. Table~\ref{tab:6} reports Reference-Match AUROC as mean $\pm$ standard deviation over three candidate-generation seeds. Policy confidence does not use world-model output; the lookahead conditions use Code2World and gWorld predictions.

\subsection{MiniWob Web Evaluation}

A web DAR evaluator is trained on Mind2Web and reused with Qwen3-VL-8B and Qwen3-VL-235B on MiniWob. We evaluate the 98 tasks expressible in the action vocabulary \{CLICK, TYPE, SELECT\}. Each task uses three environment seeds with one trajectory per seed, and all configurations share the same task--seed pairs. Table~\ref{tab:7} reports task-family results with s.e.m. over the three per-seed success rates; paired-gain s.e.m. is computed from the per-seed differences.

\subsection{Evaluator-Backbone and Objective Sensitivity}

\begin{figure}[!htbp]
\centering
\includegraphics[width=\linewidth]{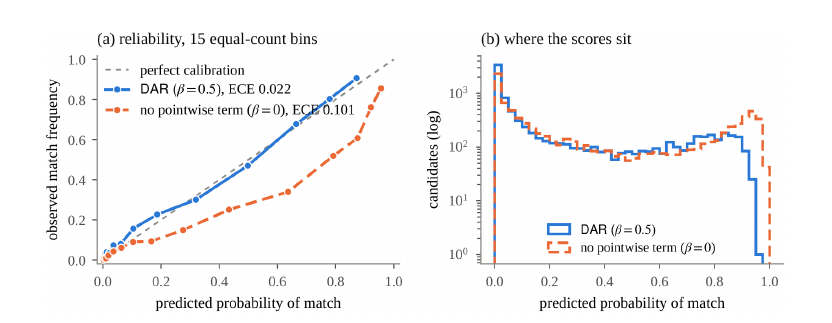}
\caption{Score behavior under pointwise supervision. Reliability curves (15 equal-count bins; 8,839 candidates) and score distributions for the full and no-pointwise Qwen3-VL-8B objectives from Table~\ref{tab:8}(b).}
\label{fig:3}
\end{figure}

Table~\ref{tab:15} adds paired online significance tests for the backbone variants in main-text Table~\ref{tab:8}(a), using the same candidate data and training configuration. Figure~\ref{fig:3} complements Table~\ref{tab:8}(b) by visualizing the score calibration and distributions of the full and no-pointwise objectives \citep{guo2017,gneiting2007}.

\begin{table}[!htbp]
\centering
\caption{Backbone sensitivity with paired online tests. AndroidControl AUROC and AndroidWorld success for the Qwen3-VL-8B proposer; online values are three-seed mean $\pm$ s.e.m. The final column reports exact paired tests against Top-1.}
\includegraphics[width=\linewidth]{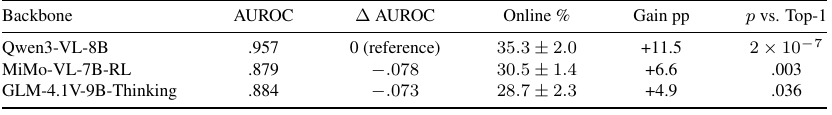}
\label{tab:15}
\end{table}

%% file: files/appendix_d.tex
\section{Prediction Diagnostics}

Table~\ref{tab:16} tests two extensions of the fixed-prediction study: longer lookahead and alternative representations of the predicted successor. All settings use Claude Opus 4.8; the one-step native-code result is the reference from Section 4.

One-step lookahead gives the largest gain in this comparison. Re-encoding the same predicted successor as an accessibility tree or plain text also reduces the gain relative to the native representation.

\begin{table}[!htbp]
\centering
\caption{Lookahead depth and representation. Reference-match change relative to Top-1, in absolute fractions, under Claude Opus 4.8. Depth results include 95\% confidence intervals; representation results report point estimates.}
\includegraphics[width=\linewidth]{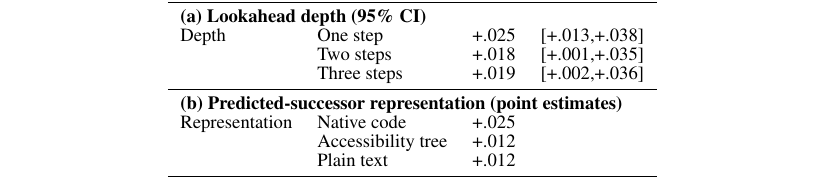}
\label{tab:16}
\end{table}

%% file: files/appendix_e.tex
\section{Qualitative Action-Selection Cases}

Among the 3,000 evaluation decisions, DAR recovers the reference action on 75 steps where Top-1 and lookahead do not. These recoveries comprise 51 clicks, 19 scrolls, 4 back actions, and one text entry. Table~\ref{tab:17} reports all 12 cases in which Top-1, lookahead, and DAR select three distinct actions; in the other 63 recoveries, Top-1 and lookahead choose equivalent actions.

\begin{table}[!htbp]
\centering
\caption{Three-way reference recoveries. All 12 decisions where Top-1, lookahead, and DAR choose distinct actions and DAR matches the reference. Parentheses show each selector's score for its selected action; DAR margin is the DAR-score difference between the reference action and Top-1.}
\includegraphics[width=\linewidth]{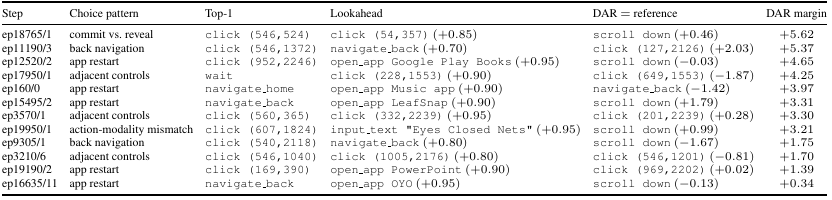}
\label{tab:17}
\end{table}

Figure~\ref{fig:2} provides one example in the main text. The three cases below highlight complementary mechanisms: ambiguity between nearby controls, variation in generated evidence for equivalent actions, and navigation away from the reference trajectory.

\begin{figure}[H]
\centering
\includegraphics[width=0.82\linewidth]{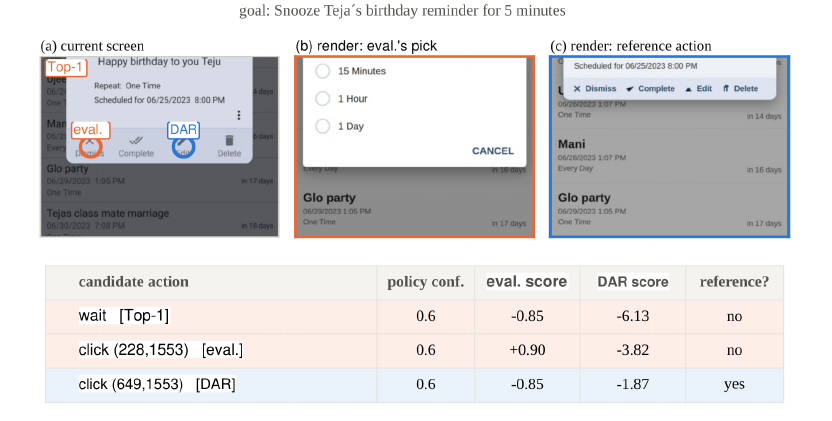}
\caption{Neighboring controls. To snooze a reminder, Top-1 waits, lookahead selects the adjacent control, and DAR selects the reference control.}
\label{fig:4}
\end{figure}

\begin{figure}[!p]
\centering
\includegraphics[width=0.78\linewidth]{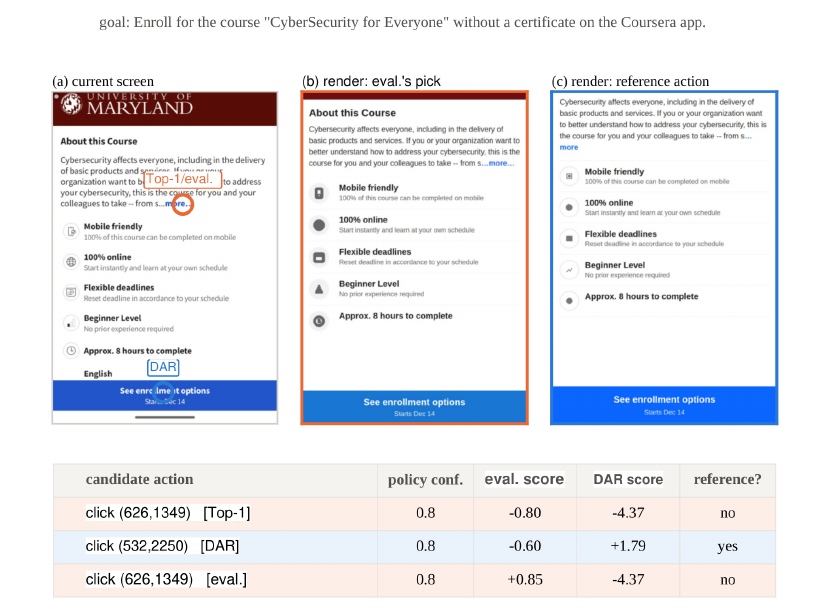}
\caption{Equivalent actions, different predictions. Equivalent non-reference clicks receive different lookahead scores; DAR scores them identically and selects the reference control.}
\label{fig:5}
\vspace{4mm}
\includegraphics[width=0.78\linewidth]{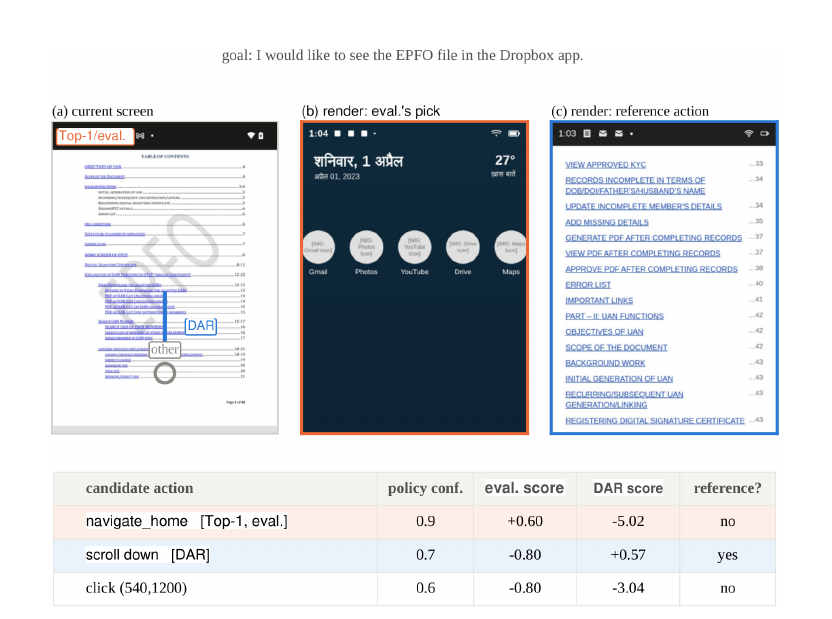}
\caption{Navigation choice. Lookahead prefers navigating home, while DAR ranks the reference scroll first.}
\label{fig:6}
\end{figure}